\documentclass[runningheads]{llncs}
\usepackage{graphicx}
\usepackage{amssymb}
\usepackage{stmaryrd}
\usepackage{subfig}
\usepackage{vcell}
\usepackage{multirow}
\usepackage{caption}
\usepackage{booktabs}
\usepackage{array}
\usepackage{ifthen}
\usepackage{algorithm}
\usepackage{algpseudocode}
\usepackage{color}
\usepackage{csquotes}
\MakeOuterQuote{"}
\usepackage{amsmath}
\usepackage{comment}
\usepackage{caption}

\DeclareMathAlphabet{\mathpzc}{OT1}{pzc}{m}{it}
\begin{document}
\title{A Capsule Network for Hierarchical Multi-Label  Image Classification
}

\author{Khondaker Tasrif Noor\inst{1}
\and
Antonio Robles-Kelly\inst{1}
\and
Brano Kusy\inst{2}
}
\authorrunning{K. T. Noor et al.}
%
\institute{Deakin University, School of IT, Waurn Ponds, VIC, Australia\\
\and
CSIRO Data 61, Queensland Centre for Advanced Technologies, QLD, Australia\\
}
\maketitle 
\begin{abstract}
Image classification is one of the most important areas in computer vision. Hierarchical multi-label classification applies when a multi-class image classification problem is arranged into smaller ones based upon a hierarchy or taxonomy. Thus, hierarchical classification modes generally provide multiple class predictions on each instance, whereby these are expected to reflect the structure of image classes as related to one another. In this paper, we propose a multi-label capsule network (ML-CapsNet) for hierarchical classification. Our ML-CapsNet predicts multiple image classes based on a hierarchical class-label tree structure. To this end, we present a loss function that takes into account the multi-label predictions of the network. As a result, the training approach for our ML-CapsNet uses a coarse to fine paradigm while maintaining consistency with the structure in the classification levels in the label-hierarchy. We also perform experiments using widely available datasets and compare the model with alternatives elsewhere in the literature. In our experiments, our ML-CapsNet yields a margin of improvement with respect to these alternative methods.


\end{abstract}

\section{Introduction}

Image classification is a classical problem in computer vision and machine learning where the aim is to recognise the image features so as to identify a target image class. Image classification has found application in areas such as face recognition \cite{wang_face_recognition:2018}, medical diagnosis \cite{zhang_mdnet:2017}, intelligent vehicles \cite{wang_intelligent_vehicles:2022} and online advertising \cite{hussain_advertising:2017} amongst others. Note that these classification tasks are often aimed at a ``flat'' class-label set where all the classes are treated equally, devoid of a taxonomical or hierarchical structure between them. This contrasts with hierarchical classification tasks, which require multiple label predictions per instance. Moreover, these should be consistent with the hierarchical structure of the class-label set under consideration. 

In this paper, we propose a multi-label image classification model (ML-CapsNet) for hierarchical image classification based on capsule networks \cite{Dynamic_CapsNet}. We note that capsule networks (CapsNets) can learn both, the image features and their transformations. This allows for a natural means to recognition by parts.  Recall that a capsule networks (CapsNets) use a set of neurons to obtain an ``activity vector''. These neurons are grouped in capsules, whereby deeper  layers account for several capsules at the preceding layer by agreement. Capsules were originally proposed in \cite{hinton2011transforming} to learn a feature of instantiation parameters which are robust to variations in position, orientation, scale and lighting. The assertion that CapsNets can overcome viewpoint invariance problems was further explored in \cite{Dynamic_CapsNet}, where the authors propose a dynamic routing procedure for routing by agreement. Hinton et al. propose in \cite{hinton:2018} a probabilistic routing approach based upon the EM-algorithm \cite{dempster:77} so as to learn part-whole relationships. Building upon the EM routing approach in \cite{hinton:2018}, Bahadori \cite{bahadori:2018} proposes a spectral method to compute the capsule activation and pose.

It is not surprising that, due to their viewpoint invariance, their capacity to address the ``Picasso effect'' in classifiers and their robustness to input perturbations when compared to other CNNs of similar size \cite{taeyoung:2019}, CapsNets have been the focus of great interest in the computer vision and machine learning communities. CapsNets have found applications in several computer vision tasks such as text classification \cite{hyperbolic_MLCaps}, 3D data processing \cite{zhao:2019}, target recognition \cite{SAR_Capsule} and image classification \cite{xiang2018ms}. They have also been used in architectures such as Siamese networks \cite{neill2018siamese}, generative adversarial networks \cite{upadhyay2018generative} and residual networks (ResNets) \cite{jampour:2021}.

Despite the interest of the community in CapsNets, to our knowledge, they have not been employed for hierarchical multi-label  classification (HMC). Since hierarchical multi-label  classification can be viewed as a generalisation of multi-class problems with subordinate, not exclusive classes, it is a challenging problem in machine learning and pattern recognition that has attracted considerable attention in the research community.  It has been tackled in a number of ways, spanning from kernel methods \cite{rousu:2006} to  decision trees \cite{vens:2008} and, more recently, artificial neural networks \cite{wehrmann:2018}. Nonetheless this attention and the fact that image HMC has been applied to the annotation of medical images \cite{dimitrovski:2011}, these methods often do not focus on images, but rather on problems such as protein structure prediction \cite{vens:2008}, data-dependent grouping \cite{ubaru:2020} or text classification \cite{meng:2019}.

Further, image hierarchical multi-label classification methods are relatively few elsewhere in the literature. This is even more surprising since it is expected that incorporating hierarchies in the model would improve generalisation, particularly when the training data is limited. Image HMC methods often employ the hierarchical semantic relationships of the target classes so as to improve visual classification results, making use of the hierarchical information as a guide to the classifier. Along these lines, word hierarchies have been applied to provide consistency across multiple datasets \cite{redmon:2017}, deliver a posterior confidence estimate \cite{davis:2021} and to optimise the accuracy-specificity trade-off in visual recognition \cite{deng:2012}. In \cite{dhall:2020}, order-preserving embeddings are used to model label-to-label and image-to-image hierarchical interactions. In \cite{B-CNN}, the authors propose a Branch-CNN (B-CNN), a CNN architecture for hierarchical data organised as a tree. The B-CNN architecture is such that the model outputs multiple predictions ordered from coarse to fine along concatenated convolutional layers. In \cite{yan:2015}, the authors propose a hierarchical deep CNN (HD-CNN) that employs component-wise pre-training with global fine-tuning making use of a multinomial logistic loss.

Here, we present a CapsNet for image HMC. Our choice of capsules is motivated by their capacity to learn relational information, seeking to profit from the capacity of CapsNets to model the semantic relationships of image features for hierarchical classification. Since CapsNets were not originally proposed for HMC tasks, we propose a modified loss function that considers the consistency of the predicted label with that endowed by the hierarchy under consideration. Further, our reconstruction loss is common to all the secondary capsule decoders, making use of the combined predictions for the hierarchies and further imposing consistency on the HMC classification task.

\begin{figure}[t]
  \fbox{\includegraphics[width=0.98\linewidth]{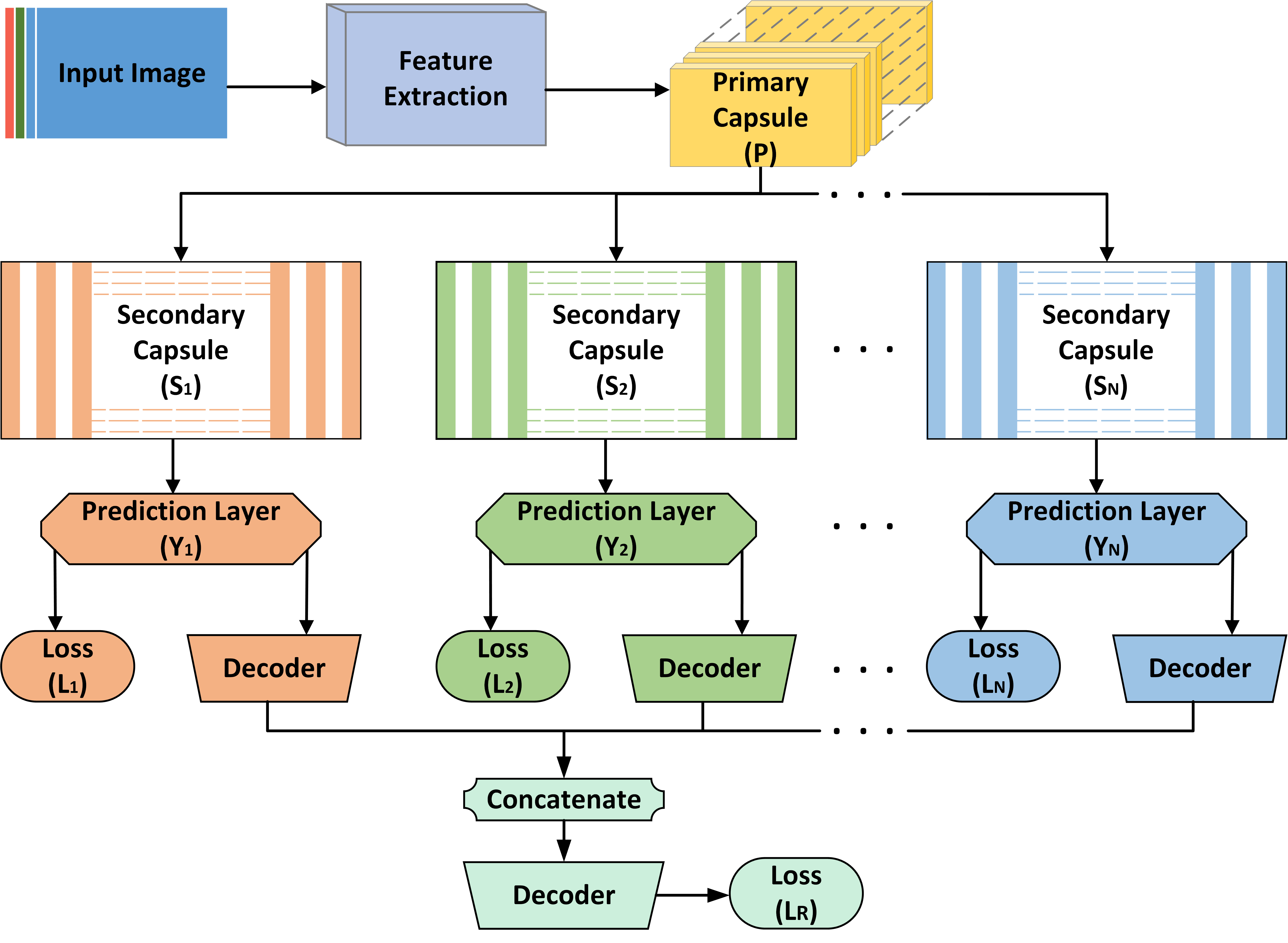}}
  \caption{Architecture of our multi-label capsule network. Our Network has multiple secondary capsules, each of these corresponding to a class-hierarchy. The output of these is used for class prediction and the reconstruction of the input image using the decoder outputs as shown in the figure.}
  \label{Fig_Arch_MLCapsNet}
\end{figure}

\section{Hierarchical Multi-label Capsules}

Our ML-CapsNet  uses a capsule architecture to predict multiple classes based on a hierarchical/taxonomical label tree. The overall architecture of ML-CapsNet is presented in Figure \ref{Fig_Arch_MLCapsNet}. Note our ML-CapsNet has a feature extraction block aimed at converting pixel intensities into local features that can be used as inputs for the primary capsule network. In our architecture, the feature map is reshaped accordingly to allow for the input features to be passed on to the primary capsule network $\mathcal{P}$, which is shared amongst all the secondary ones. These secondary capsule networks, denoted $\mathcal{S}_{n}$ in the figure, correspond to the n$^{th}$ hierarchy in the class-label set. Thus, each of the hierarchical levels, from coarse-to-fine, have their own secondary capsule networks and, therefore, there are as many of these as hierarchies. For simplicity, each of these secondary capsules sets contain $K_n$ capsules, where $K_n$ corresponds to the number of classes in the hierarchy indexed $n$. Note that, as shown in the figure, each of the secondary capsule networks $\mathcal{S}_{n}$ has their own prediction layer $\mathcal{Y}_n$, which computes the class probabilities using a logit function. Thus, the $\mathcal{Y}_{n}$ layers provide predictions for the $K_n$ classes corresponding to the $n^{th}$ hierarchy class level. 

It is also worth noting that, since CapsNets train making use of the reconstruction loss, here we use an auxiliary decoder network to reconstruct the input image. To this end, we make use of the activity vector delivered by the secondary capsule networks $\mathcal{S}_{n}$ for each hierarchy. We do this by combining the decoder outputs for each class-hierarchy into a final one via concatenation as shown in Figure \ref{Fig_Arch_MLCapsNet}. As a result, the overall loss function $L_{T}$ for our ML-CapsNet is a linear combination of all prediction losses and the reconstruction loss defined as follows
\begin{equation} 
\label{eq01}
L_{T} = \tau L_{R}+(1-\tau)\sum_{n=1}^{N}\lambda_n L_n
\end{equation}
where $n$ is the index for hierarchical level in the level tree, $N$ is the total number of levels, $\lambda_n$ a weight that moderates the contribution of each class hierarchy to the overall loss and  $\tau$ is a scalar that controls the balance between the classification loss $L_n$ and the reconstruction one $L_R$. In Equation \ref{eq01} the reconstruction loss is given by the L-2 norm between the input instance $x$ and the reconstructed one $\hat{x}$. Thus, the loss becomes
\begin{equation} 
\label{eq02}
L_{R}  = || x-\hat{x}||_2^2 \\
\end{equation}

For  ${L_n}$ we employ the hinge loss given by
\begin{equation}
\label{eq03}
L_{n}  = T_n \textrm{max}(0, m^+ - \left \|\mathbf{v}_k\right \|)^2 + \gamma (1 - T_n) \textrm{max}(0, \left \|\mathbf{v}_k\right \| - m^-)^2
\end{equation}
where $T_n=1$ if and only if the class indexed $k$ is present, $m^+$, $m^-$ and $\gamma$ are  hyper-parameters and $\mathbf{v}_k$ is the output vector for the capsule for the k$^{th}$ class under consideration.

\section{Experiments}
We have performed experiments making use of the MNIST, Fashion-MNIST, CIFAR-10 and CIFAR-100 datasets. We have also compared our results with those yielded by the B-CNN approach proposed in \cite{B-CNN} and used the CapsNet in \cite{Dynamic_CapsNet} as a baseline. Note that, nonetheless the CapsNet in \cite{Dynamic_CapsNet} is not a hierarchical classification model, it does share with our approach the routing procedure and the structure of the primary and secondary capsules. 

\newcolumntype{P}[1]{>{\centering\arraybackslash}m{#1}}
\begin{table}[!b]
\centering
\captionsetup{justification=centering}
\caption{Accuracy yielded by our ML-CapsNet, the B-CNN \cite{B-CNN} and the baseline \cite{Dynamic_CapsNet} on the  MNIST and Fashion-MNIST datasets. The absolute best are in bold.}
\label{tab-Performance-MNIST}
\begin{tabular}{P{2.2cm}|P{1.836cm}|P{1.836cm}|P{1.836cm}|P{1.836cm}|P{1.836cm}} 
\hline
\hline
\multirow{3}{*}{\textbf{Model }} & \multicolumn{5}{c}{\textbf{Accuracy (\%) }}                                                 \\ 
\cline{2-6}
                                 & \multicolumn{2}{c|}{\textbf{MNIST }} & \multicolumn{3}{c}{\textbf{Fashion-MNIST }}          \\ 
\cline{2-6}
                                 & \textbf{Coarse} & \textbf{Fine}      & \textbf{Coarse}  & \textbf{Medium} & \textbf{Fine}    \\ 
\hline
\hline
CapsNet                          & \textbf{——}     & 99.3               & \textbf{——}      & \textbf{——}     & 91.2             \\ 
\hline
B-CNN                            & 99.2            & 99.28              & 99.57            & 95.52           & 92.04            \\ 
\hline
\textbf{ML-CapsNet}              & \textbf{99.65} & \textbf{99.5}    & \textbf{99.73} & \textbf{95.93} & \textbf{92.65}  \\
\hline
\hline
\end{tabular}
\end{table}

\subsection{Implementation Details and Datasets}

We have implemented our ML-CapsNet on TensorFlow and, for all our experiments, have used the Adam optimiser with TensorFlow's default settings. For all the experiments we use exponential decay to adjust the learning rate value after every epoch, where the initial learning rate is set to $0.001$  with a decay rate of $0.995$.

As mentioned earlier, our network has a common feature extraction block and primary capsule $\mathcal{P}$ shared amongst the secondary capsules $\mathcal{S}_n$. In our implementation, the feature extraction block is comprised by two convolutional layers for the MNIST dataset and 5 convolutional layers for Fashion-MNIST, CIFAR-10 and CIFAR-100 datasets. For all the convolutional layers we have used $3\times 3$ filters with zero-padding, $ReLu$ activations and gradually increased the number of filters from $32$ filters for the first to $512$ in the following layers, i.e. $64$, $128$, $256$ and $512$. For all our results, we have used an 8-dimensional primary capsule  and all the secondary capsules $\mathcal{S}_n$ are 16-dimensional with dynamic routing \cite{Dynamic_CapsNet}. In all our experiments we have set $m^+$, $m^-$ and $\gamma$ to $0.9$, $0.1$ and $0.5$, respectively. The value of $\tau$ in Equation \ref{eq01} has been set to $0.0005$.

We have assigned coarse and medium classes to the datasets to construct a hierarchical label tree. In this setting, coarse labels are a superclass of several corresponding medium-level labels for the dataset and these labels in turn are a superclass for the fine labels prescribed for the dataset under consideration. As a result, each instance will have multiple labels in the hierarchy, i.e. one per level. Recall that the MNIST and Fashion-MNIST dataset contains $28\times28$ grey-scale images. Both datasets have $60,000$ training and $10,000$ testing images. For the MNIST, which contains images of handwritten digits, we have followed \cite{B-CNN} and added five coarse classes for the dataset. The Fashion-MNIST dataset is similar to the MNIST dataset and contains images of fashion products for ten fine classes. We use the coarse to medium hierarchy presented in \cite{CNN_fashionimage}, which manually adds a coarse and medium level for the dataset. In \cite{B-CNN}, the coarse level has two classes, and the medium level has six classes. The CIFAR-10 and CIFAR-100 dataset consist $32\times32$ colour images and have $50,000$ training images and $10,000$ testing images in 10 and 100 fine classes, respectively. Here, we use the class-hierarchy used in \cite{B-CNN}, which adds an additional coarse and medium levels for the dataset whereby the coarse level contains two classes and the medium contains seven classes. The CIFAR-100 dataset is similar to CIFAR-10, except it has 100 fine and 20 medium classes. We have used the hierarchy in \cite{B-CNN}, which groups the medium and fine classes into eight coarse ones. 

\begin{table}[!t]
\centering
\captionsetup{justification=centering}
\caption{Accuracy yielded by our ML-CapsNet, the B-CNN \cite{B-CNN}  and the baseline \cite{Dynamic_CapsNet} on the CIFAR-10 and CIFAR-100 datasets. The absolute best are in bold.}
\label{tab-Performance-CIFAR}
\begin{tabular}{P{2.2cm}|P{1.53cm}|P{1.53cm}|P{1.53cm}|P{1.53cm}|P{1.53cm}|P{1.53cm}} 
\hline
\hline
\multirow{3}{*}{\textbf{Model }} & \multicolumn{6}{c}{\textbf{Accuracy (\%) }}                                                                         \\ 
\cline{2-7}
                                 & \multicolumn{3}{c|}{\textbf{CIFAR-10}}                   & \multicolumn{3}{c}{\textbf{CIFAR-100 }}                  \\ 
\cline{2-7}
                                 & \textbf{ Coarse } & \textbf{ Medium } & \textbf{ Fine }  & \textbf{ Coarse } & \textbf{ Medium } & \textbf{ Fine }   \\ 
\hline
\hline
CapsNet                          & \textbf{——}                & \textbf{——}                & 70.42            & \textbf{——}                & \textbf{——}                & 34.93             \\ 
\hline
B-CNN                            & 95.63             & 86.95             & 84.95            & 71.04             & 61.94             & 55.52             \\ 
\hline
\textbf{ML-CapsNet}             & \textbf{97.52}  & \textbf{89.27}  & \textbf{85.72} & \textbf{78.73}  & \textbf{70.15}  & \textbf{60.18}  \\
\hline
\hline
\end{tabular}
\end{table}

\subsection{Experimental Setup}

As mentioned earlier, in order to compare our ML-CapsNet with alternatives elsewhere in the literature, we have used the CapsNet as originally proposed in \cite{Dynamic_CapsNet} as a baseline and compared our results with those yielded by the B-CNN \cite{B-CNN}. In all our experiments, we normalise the training and testing data by subtracting the mean and dividing by the standard deviation. For all the datasets, the B-CNN follows the exact model architecture and training parameters used by the authors in \cite{B-CNN}. For training the B-CNN model on the Fashion-MNIST dataset, we followed the architecture for the MNIST dataset and added an additional branch for the fine level. 

When training our ML-CapsNet on the CIFAR-10 and CIFAR-100 datasets, we apply MixUp data augmentation \cite{zhang2018mixup} with $\alpha = 0.2$. Also, recall that the $\lambda_n$ values in Equation \ref{eq01} govern the contribution of each hierarchy level to the overall loss function. Thus, for our ML-CapsNet, we  adjust the $\lambda_{n}$ values as the training progresses so as to shift the importance of the class-level hierarchy from course-to-fine. As a result, on the MNIST dataset we set the initial $\lambda_{n}$ to $0.90, 0.10$ for the coarse and fine levels, respectively. The value of $\lambda_{n}$ is then set to $0.10, 0.90$ after $5$ epochs and $0.02 0.98$ after $10$. For the Fashion-MNIST dataset the initial loss weight values for $\lambda_{n}$ are $0.98, 0.01, 0.01$ for the coarse, medium and fine hierarchies. The value of $\lambda_{n}$ after 5 epochs becomes 0.10, 0.70, 0.20, after 10 epochs shifts to 0.07, 0.10, 0.83, at 15 epochs is set to 0.05, 0.05, 0.90 and, finally, at 25 epochs assumes the value of 0.01, 0.01, 0.98. For the CIFAR-10 dataset, the initial $\lambda_{n}$ values are $0.90$, $0.05$ and $0.05$ for the coarse, medium  and fine class-hierarchy levels. The $\lambda_{n}$ then changes to $0.10$, $0.70$ and $0.20$ after Epoch 5 and, at Epoch 11 is set to $0.07$, $0.20$ and $0.73$. At Epoch 17 is set to $0.05$, $0.15$ and $0.80$, taking its final value of  $0.05$, $0.10$ and $0.85$ at Epoch 24. For the CIFAR-100 dataset the initial values are $0.90$, $0.08$ and $0.02$, which then change to $0.20$, $0.70$ and $0.10$ after 7 epochs. After Epoch 15 we set these to $0.15$, $0.30$ and $0.55$. At Epoch 22 we modify them to be $0.10$, $0.15$ and $0.75$ and, at Epoch 33, they take their final value of $0.05$, $0.15$ and $0.80$.

\begin{figure}[!t]
	\centering
	\subfloat[MNIST \label{Fig_Comp_MNIST}]{{\fbox{\includegraphics[width=0.475\linewidth, height=5cm]{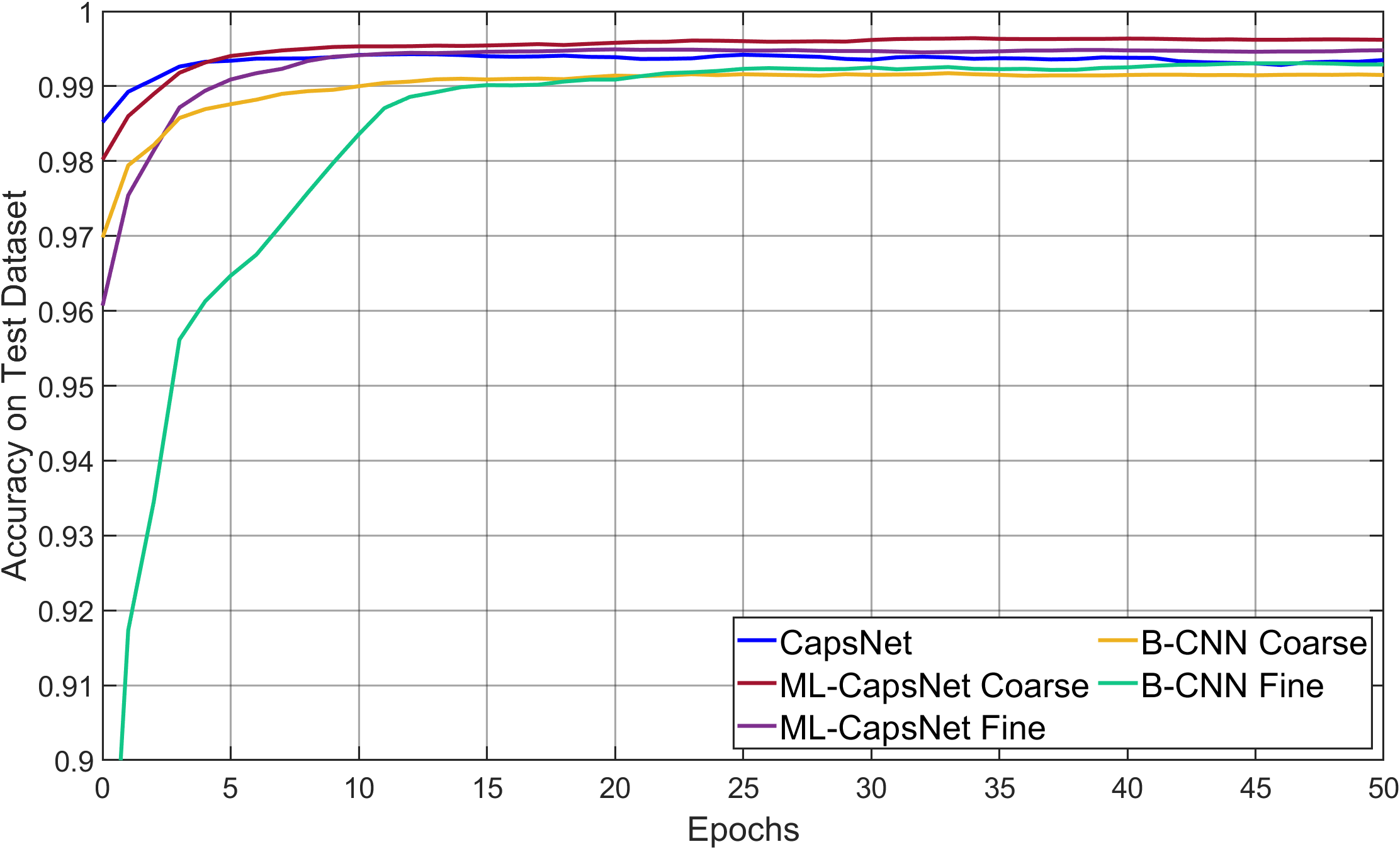}}}}%
	\thinspace
	\subfloat[Fashion-MNIST \label{Fig_Comp_FMNIST}]{{\fbox{\includegraphics[width=0.475\linewidth, height=5cm]{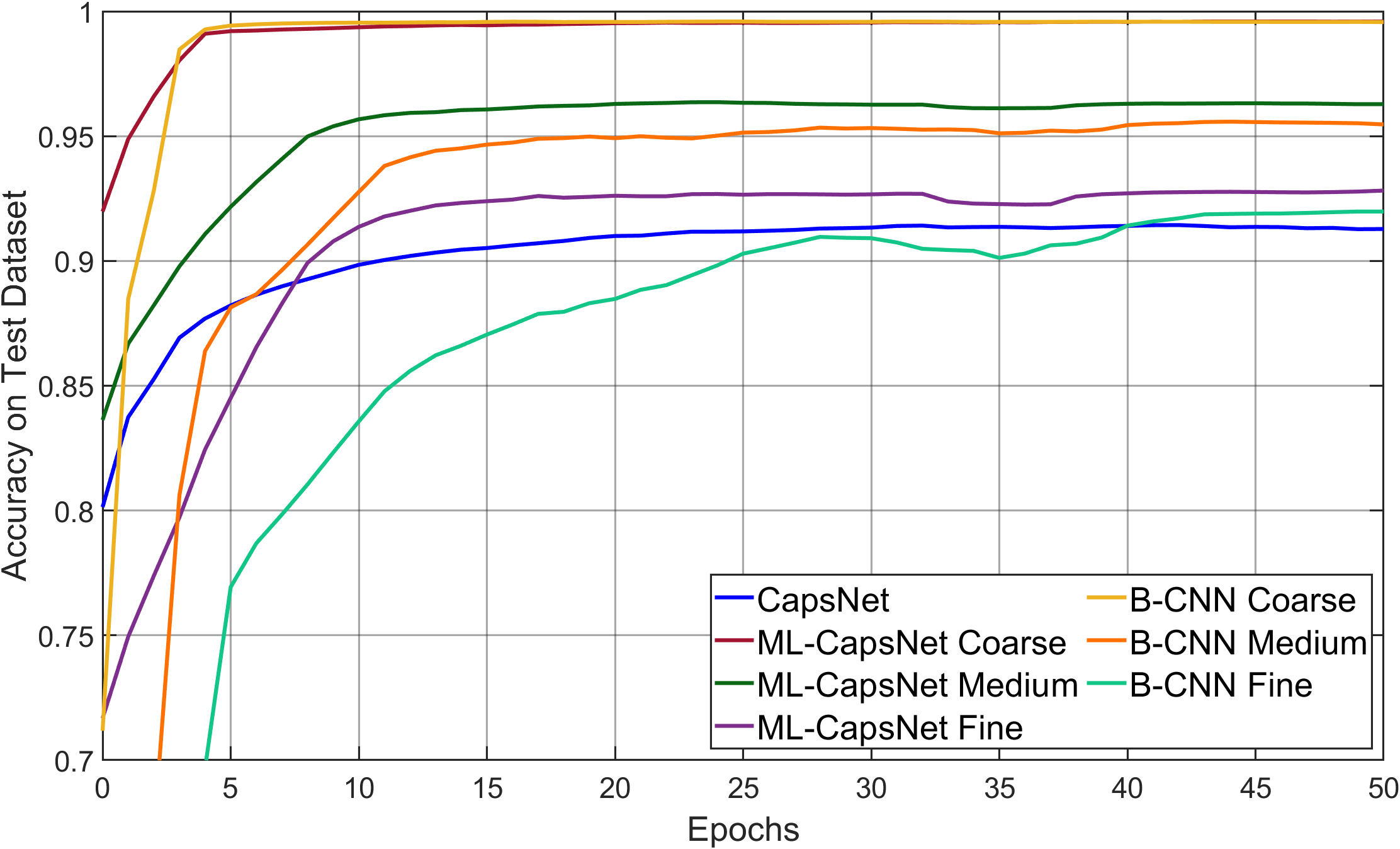}}}}%
	\caption{Accuracy as a function of training epoch for all the models under consideration. The left-hand panel shows the plots for the MNIST dataset whereas the right-hand panel corresponds to the Fashion-MNIST dataset.}%
	\label{Fig_Comp_AllMNIST}%
\end{figure}
\begin{figure}[!t]
	\centering
	\subfloat[CIFAR-10 \label{Fig_Comp_CIFAR10}]{{\fbox{\includegraphics[width=0.475\linewidth, height=5cm]{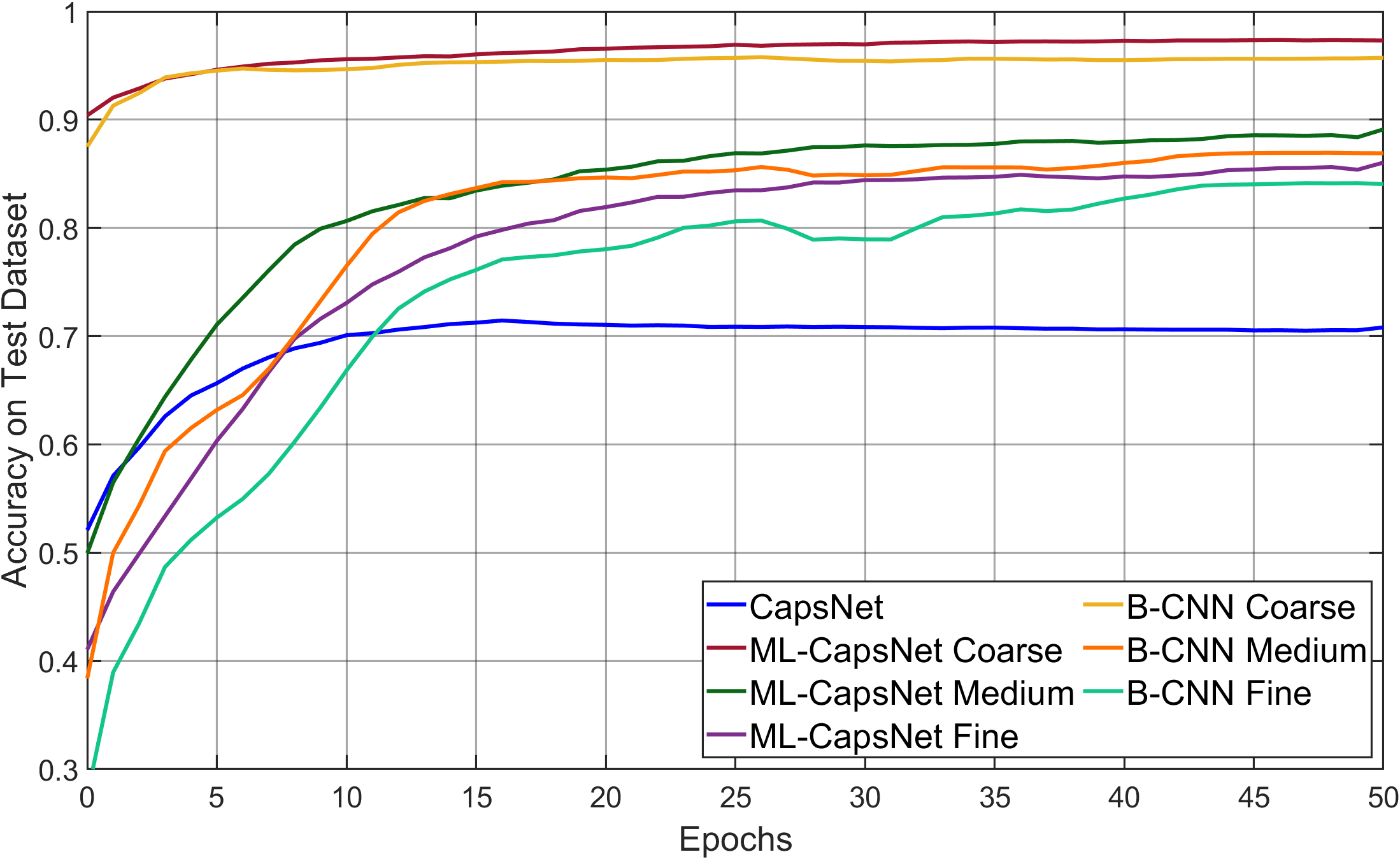}}}}%
	\thinspace
	\subfloat[CIFAR-100 \label{Fig_Comp_CIFAR100}]{{\fbox{\includegraphics[width=0.475\linewidth, height=5cm]{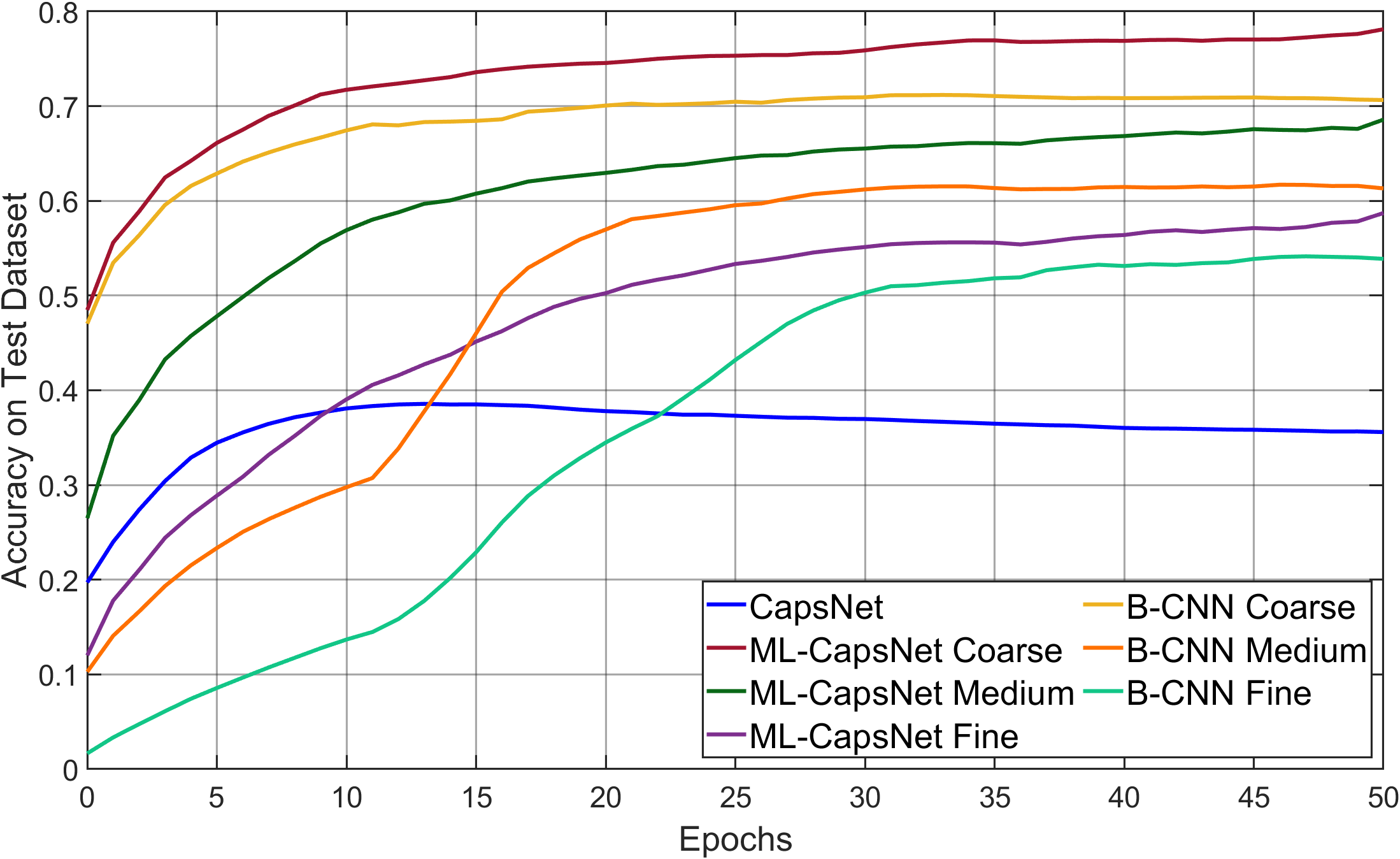}}}}%
	\caption{Accuracy as a function of training epoch for all the models under consideration. The left-hand panel shows the plots for the CIFAR-10  dataset whereas the right-hand panel corresponds to the CIFAR-100 dataset.}%
	\label{Fig_Comp_AllCIFAR}%
\end{figure}

\subsection{Results}\label{sec_result}

We now turn our attention to the results yielded by our network, the CapsNet as originally proposed in \cite{Dynamic_CapsNet} and the B-CNN in \cite{B-CNN} when applied to the four datasets under consideration. In Table \ref{tab-Performance-MNIST} we show the performance yielded by these when applied to the MNIST and Fashion-MNIST dataset. Similarly, in Table \ref{tab-Performance-CIFAR}, we show the accuracy for our network and the alternatives when applied to the CIFAR-10 and CIFAR-100 datasets. In the tables, we show the accuracy for all the levels of the applicable class-hierarchies for both, our network and the B-CNN. We also show the performance on the CapsNet in \cite{Dynamic_CapsNet} for the fine class-hierarchy. We do this since the baseline is not a hierarchical classification one, rather the capsule network as originally proposed in \cite{Dynamic_CapsNet} and, therefore, the medium and coarse label hierarchies do not apply. 

Note that, for all our experiments, our network outperforms the alternatives. It is also worth noting that, as compared with the baseline, it fairs much better for the more complex datasets of CIFAR-10 and CIFAR-100. This is also consistent with the plots in Figures \ref{Fig_Comp_AllMNIST} and \ref{Fig_Comp_AllCIFAR}, which show the model accuracy on the dataset under consideration for every one of their corresponding class-hierarchies. Note that in several cases, our network not only outperforms B-CNN, but also converges faster. It is also important to note that, as compared to the baseline, our ML-CapsNet employs the hierarchical structure of the class-set to impose consistency through the loss. This appears to introduce a structural constraint on the prediction that helps improve the results. 

\begin{figure}[!b]
  \includegraphics[width=\linewidth]{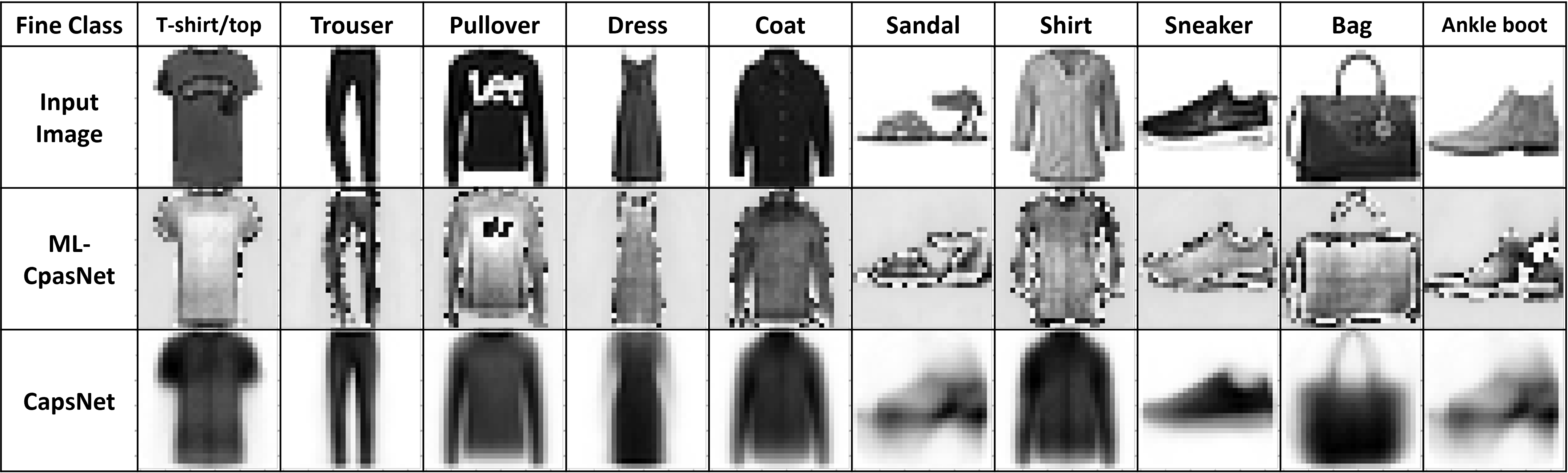}
  \caption{Sample reconstructed images for the Fashion-MNIST dataset. From top to bottom we show the input image, the image reconstructed by our ML-Capsnet and that reconstructed using the capsule network in \cite{Dynamic_CapsNet}.}
  \label{Fig_Reconstruction_FMNIST}
\end{figure}
This is consistent with the reconstruction results shown in Figure \ref{Fig_Reconstruction_FMNIST}. In the figure, we show the reconstruction results for the classes in the Fashion-MNIST for both, our ML-CapsNet and the capsule network as originally proposed in \cite{Dynamic_CapsNet}. In the figure, from top-to-bottom, the rows show the input image, the image reconstructed by our network and that yielded by that in \cite{Dynamic_CapsNet}. Note the images reconstructed using our network are much sharper, showing better detail and being much less blurred. This is somewhat expected, since the reconstruction error is also used by the loss and, hence, better reconstruction should yield higher accuracy and vice versa. 

\section{Conclusions}
In this paper, we have presented a capsule network for image classification, which uses capsules to predict multiple hierarchical classes. The network presented here, which we name ML-CapsNet, employs a shared primary capsule, making use of a secondary one for each class-label set. To enforce the multi-label structure into the classification task, we employ a loss which balances the contribution of each of the class-sets. The loss proposed here not only enforces consistency with the label structure, but incorporates the reconstruction loss making use of a common encoder. We have shown results on four separate widely available datasets. In our experiments, our ML-CapsNet outperforms the B-CNN \cite{B-CNN} and the classical capsule network in \cite{Dynamic_CapsNet}. As expected, it also delivers better reconstructed images than those yielded by the network in \cite{Dynamic_CapsNet}.

\bibliographystyle{splncs04}
\bibliography{References_CNMLC}

\end{document}